\documentclass[sigconf]{acmart}

\usepackage{outlines}
\usepackage{todonotes}
\usepackage{xcolor}
\usepackage[mathscr]{eucal}
\usepackage{bm}
\usepackage{algpseudocode} 
\usepackage{algorithm}
\usepackage{xifthen}
\usepackage[export]{adjustbox}

\usepackage[htt]{hyphenat}

\NewDocumentCommand{\vect}{ O{} O{} m }{\mathbf{#3}\ifthenelse{\isempty{#1}}{}{^{(#1)}}\ifthenelse{\isempty{#2}}{}{_{#2}}}

\NewDocumentCommand{\mat}{ O{} O{} m }{\mathbf{#3}\ifthenelse{\isempty{#1}}{}{^{(#1)}}\ifthenelse{\isempty{#2}}{}{_{#2}}}

\NewDocumentCommand{\ten}{ O{} O{} m }{\pmb{\mathscr{#3}}\ifthenelse{\isempty{#1}}{}{^{(#1)}}\ifthenelse{\isempty{#2}}{}{_{#2}}}

\AtBeginDocument{%
  }

\setcopyright{acmlicensed}
\copyrightyear{20xx}
\acmYear{20xx}
\acmDOI{XXXXXXX.XXXXXXX}

\acmConference[DocEng '24]{The 24th ACM Symposium on Document Engineering}{August 20--23,
  2024}{San Jose, CA, USA}
\acmISBN{978-1-4503-XXXX-X/18/06}

\begin{document}

\title{TopicTag: Automatic Annotation of NMF Topic Models Using Chain of Thought and  Prompt Tuning with LLMs}

\author{Selma Wanna}
\author{Nicholas Solovyev}
\affiliation{
    \institution{Advanced Research in Cyber Systems}
    \institution{Theoretical Division\\ Los Alamos National Laboratory}
    \country{USA}
}

\author{Ryan Barron}
\author{Maksim E. Eren}
\affiliation{
    \institution{Theoretical Division}
    \institution{Advanced Research in Cyber Systems\\Los Alamos National Laboratory}
    \country{USA}
}

\author{Manish Bhattarai}
\author{Kim {\O}. Rasmussen}
\author{Boian S. Alexandrov}
\affiliation{
    \institution{Theoretical Division\\ Los Alamos National Laboratory}
    \country{USA}
}

\renewcommand{\shortauthors}{Wanna et al.}

\begin{abstract}
Topic modeling is a technique for organizing and extracting themes from large collections of unstructured text. Non-negative matrix factorization (NMF) is a common unsupervised approach that decomposes a term frequency-inverse document frequency (TF-IDF) matrix to uncover latent topics and segment the dataset accordingly. While useful for highlighting patterns and clustering documents, NMF does not provide explicit topic labels, necessitating subject matter experts (SMEs) to assign labels manually. We present a methodology for automating topic labeling in documents clustered via NMF with automatic model determination (NMFk). By leveraging the output of NMFk and employing prompt engineering, we utilize large language models (LLMs) to generate accurate topic labels. Our case study on over 34,000 scientific abstracts on Knowledge Graphs demonstrates the effectiveness of our method in enhancing knowledge management and document organization.

\end{abstract}




\maketitle

\section{Introduction}
\label{sec:introduction}



The rapid growth of digital text data has necessitated the development of advanced techniques for organizing, cataloging, and extracting information from large datasets. Topic modeling, a powerful document engineering technique, has been widely used to segment large datasets into manageable clusters of related documents. Another critical task in document organization is assigning labels that summarize the themes of these topics. Traditionally, this has been done by subject-matter experts (SMEs) through a manual and time-consuming investigation of each topic. In this work, we introduce a novel, automated topic labeling technique that leverages patterns obtained through dimensionality reduction for prompt-tuning, utilizing Large Language Models (LLMs).

One common approach to topic modeling is through non-negative matrix factorization (NMF) of the term frequency-inverse document frequency (TF-IDF) matrix. Given a TF-IDF matrix, $\mat{X} \in \rm I\!R_{+}^{m \times n}$ where $m$ is the number of tokens in the vocabulary and $n$ is the number of documents in the corpus, $\mat{X}$ can be approximated as the product of two non-negative matrices $\mat{W} \in \rm I\!R_{+}^{m \times k}$ and $\mat{H} \in \rm I\!R_{+}^{k \times n}$, such that $\mat{X}_{ij} \approx \sum^{k}_{s} \mat{W}_{is} \mat{H}_{sj}$, and the low-rank $k\ll n,m$. Here $k$, the matrix rank, is the number of topics, the rows of $\mat{W}$ represent the distribution of the tokens into each $k$ topics, and the columns of $\mat{H}$ are the coordinates of the documents in the latent topic space (i.e. clustering of the documents into $k$ topics). The careful selection of $k$ is important as too small value of $k$ can result on poor topic separation (\textit{under-fitting}), and too large a value of $k$ will result in noisy topics (\textit{over-fitting}). In this work, we use NMF with automatic model determination (NMFk)\footnote{NMFk is available in \url{https://github.com/lanl/T-ELF}.} to heuristically estimate the number of topics ($k$) \cite{eren2022, TELF}. Despite its effectiveness, NMFk does not inherently provide labels for the discovered topics, necessitating the involvement of SMEs to review the latent factors, clustered documents, and other derivatives of the factorization such as word-clouds to assign meaningful topic labels. 

\begin{figure}
    \centering
    \resizebox*{8.5cm}{!}{\includegraphics[width=\textwidth]{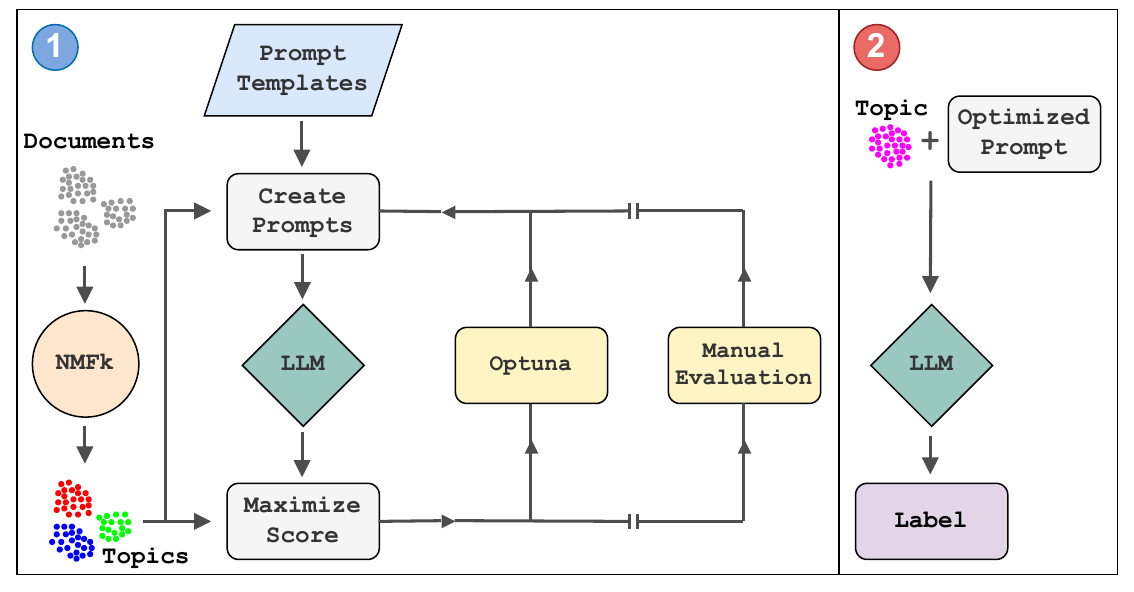}}
        \caption{
The TopicTag pipeline. Stage 1 (left) illustrates the prompt-optimization framework applied to our training set of topic clusters. Documents are processed by the NMFk algorithm to generate feature information, which is then integrated into prompts. These prompts are evaluated by LLMs, with their label predictions compared against ground truth labels. Prompts are refined by maximizing NLG or human rater scores. In Stage 2, the optimal prompts are assessed on the test set.}
    \label{fig:method-summary}
\vspace*{-\baselineskip}
\end{figure}

Our approach uniquely integrates the output of an NMFk decomposition with prompt engineering, chain of thought prompting, and Optuna \cite{akiba2019optuna} for prompt tuning to generate accurate and descriptive topic labels, significantly reducing the need for SME intervention. This paper details our methodology and demonstrates its efficacy through a case study on over 34,000 abstracts of scientific literature on Knowledge Graphs (KG). The results highlight the potential of our method to enhance knowledge management and information discovery, making it a valuable tool for organizing and understanding large text datasets across various domains.

\section{Relevant Work}
\label{sec:relevant_work}

The outputs of topic modeling are typically labeled by an SME using some top-ranked vocabulary based on the conditional probability \( p(w_i \mid k) \) of a token  $w_i$ appearing in a topic $k$ \cite{blei2003}. This approach is both exhausting and prone to error. It demands that an SME in the target domain quickly analyze a topic cluster's top vocabulary and produce a label. Even with a highly experienced SME, the process is tedious and is susceptible to subjectivity. To this end, automatic topic label generation has been a long standing problem.

One common method of creating topic labels is to extract many candidate labels from inputs such as documents titles, frequent phrases, and the top vocabulary and then select the best candidate by maximizing mutual information between a label and the topic model \cite{mei2007, kou2015, basave2014}. Other approaches follow a similar extractive candidate selection and evaluation paradigm but incorporate Information Retrieval to enrich the source of potential candiate labels \cite{giorgi2023open}. More recently, abstractive summarization tasks using deep learning techniques have attracted considerable research attention \cite{zakkas2024, congbo2020, ghadimi2022}. These methods can provide detailed summaries that offer more comprehensive insights into topics but come with certain trade-offs. Generating and verifying the relevance and accuracy of long-form summaries for the multiple documents composing a topic poses a significant challenge \cite{giorgi2023open}. Conversely, short labels often generalize better and provide a quick, at-a-glance overview of the topic. In the rest of this paper we show how the latent features from NMFk topic modeling can be used with prompt engineering to produce high quality topic labels.

\section{Methods}
\label{sec:methods}
In this section, we define the labeling task and outline the construction of the TopicTag pipeline, as illustrated in Figure \ref{fig:method-summary}.





\subsection{Task Definition}
Several criterion must be met for a topic label to be effective. This criterion can be summed up as a maximization of relevance, coverage, and discrimination for topic label \( t \in T \) assigned to a topic cluster \( k \in K \) \cite{wan2016}. High relevance is essential; the label must be semantically aligned with the topic and closely related to all representative documents. Secondly, high coverage requires that the label should encapsulate as much semantic information about the topic as possible and should at a high level be applicable to all documents in the topic. Additionally, when creating a set of labels for all topics identified in a document collection, high discrimination is necessary. Labels for different topics must exhibit inter-topic discrimination to help users understand each topic. If labels for multiple topics are too similar, users may find it difficult to distinguish between them. Therefore, the higher the inter-topic discrimination, the better the labels.

\subsection{Prompt} 
In our experiments, we focus on developing zero-shot prompts, comprising a system and task instruction for the LLM. To mitigate the high context window space required by few-shot examples, we employ various Chain-of-Thought templates~\cite{wei2022chain}. 

\subsection{Models}
\label{sec:models}

We evaluate four LLMs in this study. Initially, \texttt{Mistral-7B-\\Instruct-v0.2}~\cite{jiang2023mistral} was employed for efficient prompt engineering and optimization, based on the premise that effective prompts would generalize across varying model scales and architectures. Subsequently, the optimized prompt derived from \texttt{Mistral-7B-Instruct-v0.2} was used to initiate the prompt engineering process for four other LLMs: \texttt{Mistral-7B-Instruct-v0.3}, \texttt{Meta-\\Llama-3-8B-Instruct}~\cite{llama3modelcard}, \texttt{Meta-Llama-3-70B-Instruct}~\cite{llama3modelcard}, and \texttt{gpt-4o}.

\subsection{Metrics}
For each topic model cluster, we developed manually assigned SME labels to serve as ground truths. During optimization, to assess the quality of the LLM responses, we utilized several Natural Language Generation (NLG) metrics, including BERTScore~\cite{Zhang2020BERTScore}, BLEU~\cite{papineni-etal-2002-bleu}, and ROUGE~\cite{lin-2004-rouge}. For the second stage of LLM filtering, responses were qualitatively evaluated using a 3-point scoring system based on the concept of "relevance." This concept, as defined by text summarization researchers~\cite{ernst-etal-2023-examining}, examines whether the most salient concepts from the text are accurately represented. For our final evaluation, we employ a 5-point version of this metric.

\subsection{Prompt Filtering as Prompt Engineering}

Figure \ref{fig:method-summary} provides an overview of our TopicTag pipeline. The initial stage involves processing the output of our NMFk model. Each $n$ document is assigned a topic via a column-wise \texttt{argmax} operation on $\mat{H}$. Since the columns of $\mat{H}$ are the coordinates of the documents in the latent topic space, this is equivalent to a soft clustering of our documents, and the proximity of each coordinate to the cluster centroid can be further used for ranking documents within the topic \cite{vangara2021finding}. Similarly, $\mat{W}$ can be processed to extract the top \textit{m} tokens for each cluster. This post-processing then yields the top titles, abstracts, keywords, n-grams for a given topic. These features are assessed for their suitability in the topic labeling task by integrating them into various Chain-of-Thought prompt templates~\cite{wei2022chain}.


To identify the most effective features, we employ a two-step filtering algorithm that compares the responses of an LLM against ground-truth labels assigned to each topic model cluster.

In the first step, an automated coarse-grain search is conducted using the Tree-structured Parzen Estimator (TPE) sampling algorithm from \texttt{Optuna}~\cite{akiba2019optuna}. This algorithm explores the document feature space (such as top SME keywords, top words, top n-grams, top titles, document coordinates, TF-IDF weights, etc.), obtained from NMFk, and hyperparameter configurations of \texttt{Mistral-7B-Instruct-v0.2} to maximize BERTScore. This step typically involves 2-3 iterations to find features that significantly reduce the BERTScore, allowing us to narrow our search space.

The second step involves qualitative filtering through SME preference ratings. Evaluators categorize responses as bad, okay, or good. Good responses are further analyzed to identify commonalities in prompting features and LLM generation hyperparameters. This step is repeated 3-4 times to refine the prompts, which are then used to guide the search for additional LLM prompts (see Sec. \ref{sec:models}). The qualitative filtering process is repeated an additional 1-4 times to ensure the selection of ideal prompts for the new models.

\section{Results}
\label{sec:results}
This section details the experimental setups and results of two studies. The first study examines the correlation between human rater agreement and traditional NLG metrics. The second study investigates the factors contributing to more successful prompts and identifies the best-performing LLM and prompt combination.

The experimental dataset and its decomposition come from a previously collected set of documents for another work that focuses on knowledge graphs and knowledge base management. For both studies, we implemented a train/test split on the 28 topics generated by the NMFk. The training set, on which we applied our TopicTag pipeline, comprises the first seven topics produced by NMFk. The remaining 21 topics serve as the evaluation set.

Given the extensive output space encompassing document features, prompt templates, and LLM hyperparameter configurations, we generated approximately 840 LLM outputs. From these, we randomly sampled roughly 160 responses for manual analysis.


\subsection{Experiment 1: NLG Metric Study}
\label{sec:experiment_1}
In this study, we enlisted seven raters (authors of this paper) to score the outputs of LLMs and measure the correlation between NLG metrics and human preferences. Scores were reported on a 5-point scale. Table \ref{tab:CORR-STUDY} summarizes these results. Overall, the studied metrics do not exhibit strong correlations with human judgments. Due to its weak signal, we have omitted the BLEU score from our results. ROUGE-L shows the highest correlation with human judgments; however, an average $r^2=0.139$ still indicates poor performance. Given our scores are not correlated with human SME raters, we employ these metrics solely to filter out egregiously incorrect responses. This highlights the complexity of evaluating topic labels that may not be prominent in a cluster's vocabulary. Unlike translation and summarization, current NLG metrics fail to capture the trade-offs between granularity and broadness in labeling. Therefore, there is a need for more robust alternatives that can capture the nuances of SME labeling in future work.



\begin{table}
\small
  \caption{We report the Inter-Annotator Agreement (IAA) as Cohen's Kappa. Our scores are consistent with "fair" agreement. We also report the $r^2$ values between human rater scores and NLG metrics: BERTScore and ROUGE-L.}
  \label{tab:CORR-STUDY}
  \begin{tabular}{cccl}
    \toprule
    Annotators & IAA & $r^2$: BERTScore & $r^2$:  ROUGE-L \\
    \midrule
    Group 1 & 0.459 & 0.037 & 0.112\\
    Group 2 & 0.419 & 0.081 & 0.204\\
    Group 3 & 0.369 & 0.156 & 0.127\\
    Group 4 & 0.425 & 0.037 & 0.112\\
  \bottomrule
\end{tabular}
\end{table}

Despite ROUGE-L's better performance, its n-gram nature does not provide a smooth score suitable for the TPE optimizer to effectively search over. Therefore, we leverage BERTScore at the initial stage of our optimization to eliminate document information that results in substantially poor prompts. In practice, features frequently filtered out included TF-IDF scores on keyword information, bi- and tri-gram information, and $\mat{H}$ coordinates distances between a document and their topic's centroid. 
However, ordering abstracts and title information based on distance to the centroid did enhance LLM generation quality.

\subsection{Experiment 2: Evaluating LLM Topic Labels}

\begin{table*}[h!]
\small
  \caption{Qualitative results comparing LLM label predictions against SME ground truths.}
  \label{tab:QualStudy}
  \begin{tabular}{cccc}
    \toprule
    Topic \# & SME Rating & Ground Truth & \texttt{Llama-3-8B-Instruct} \\
    \midrule
     9 & 5 & "Domain Ontology Construction" & "Ontology Construction Management and Extraction" \\
     8 & 5 & "Natural Language Processing" & "Natural Language Processing and Formal Language Theory" \\
     13 & 4 & "Graph Embeddings" & "Graph Neural Networks for Node Classification"\\
     7 & 3 & "Dependency Relation Extraction" & "Knowledge Graph Completion and Relation Extraction" \\
     10 & 2 & "Applications and Use Cases of GNNs" & "Machine Learning Applications in Science" \\
     26 & 2 & "Knowledge Base Construction" & "Construction Project Management" \\  
     15 & 1 & "Semantic Knowledge Representation" & "Chinese Lexical Semantics"\\
  \bottomrule
\end{tabular}
\vspace{-0.5em}
\end{table*}

\begin{figure}
    \centering
    \resizebox*{8.5cm}{!}{\includegraphics[width=\textwidth]{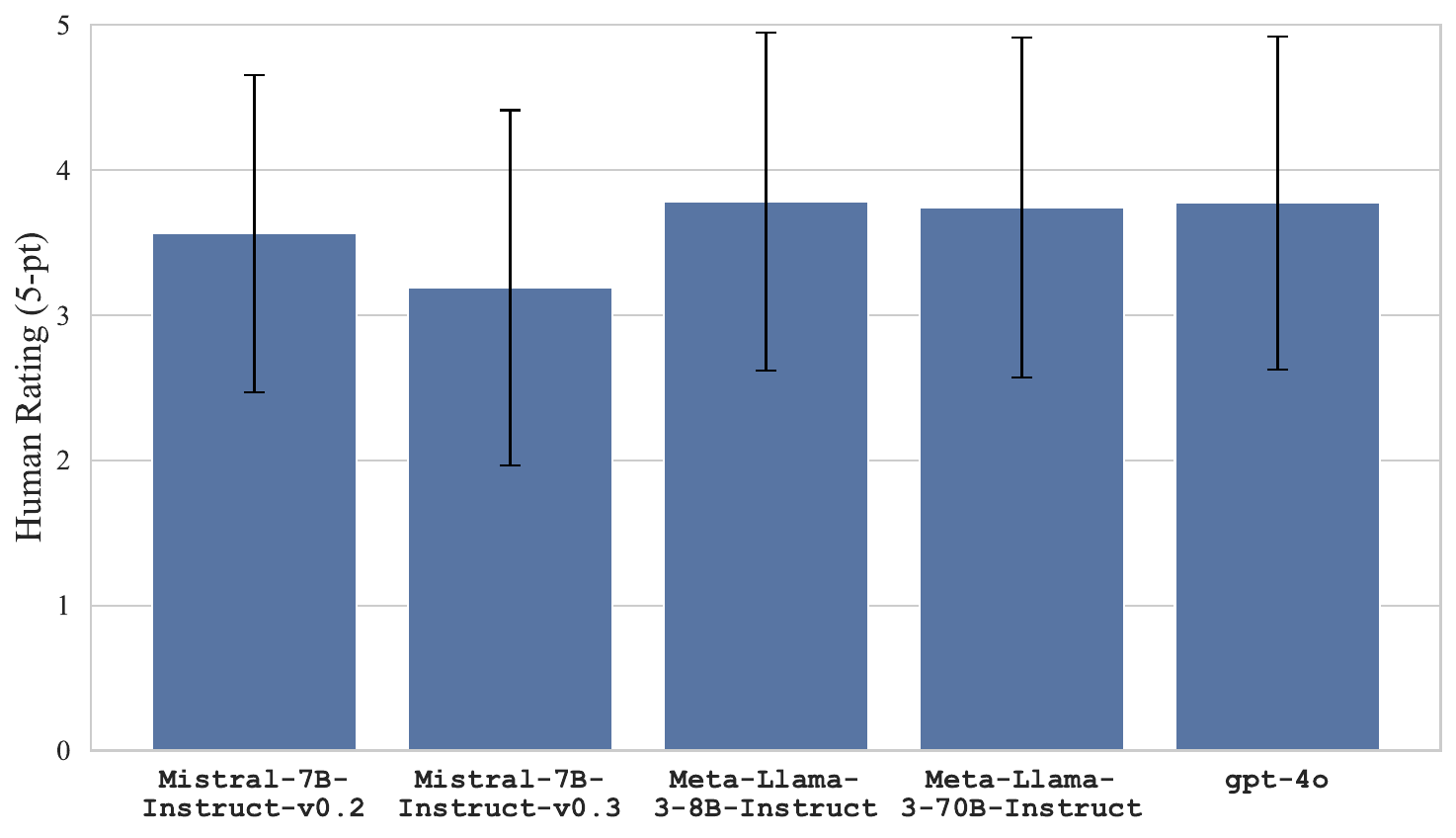}}
        \caption{The average performance for each tested LLM marginalized over our test set, prompt templates, document features, and LLM hyperparameter configurations. We believe these variations in conjunction with the reported IAA in Table \ref{tab:CORR-STUDY} lead to the reported variance. }
    \label{fig:barchart}
\vspace*{-\baselineskip}
\end{figure}


\label{sec:experiment_2}


This study evaluates the efficacy of the TopicTag pipeline using human rater data collected in Section \ref{sec:experiment_1} and summarized in Figure \ref{fig:barchart}. The results demonstrate that our approach generalizes well across LLMs of varying architecture and size. Notably, \texttt{Meta-Llama-3-8B-Instruct} was the top-performing model with an average score of 3.78. Examples of \texttt{Meta-Llama-3-8B-Instruct}'s predicted topic labels are reported in Table \ref{tab:QualStudy}. \texttt{Mistral-7B-Instruct-v0.3} had the lowest performance with an average score of 3.19. However, the difference between these averages is 0.59, which is less than a single point on our 5-point scale, perhaps indicating a relatively minor distinction. An important consideration is the relative optimization speed of smaller models compared to larger ones. Due to time and resource constraints, we may have undertuned \texttt{Meta-Llama-3-70B-Instruct} and \texttt{gpt-4o}, applying only 1-2 optimization steps, whereas smaller models underwent 3-4 optimization steps. This difference in optimization effort may have impacted the performance outcomes of the larger models.

From our analysis, we determined that using the following CoT prompt template: \texttt{"You are a document understander. Using your expertise, label this topic cluster by thinking step-by-step:\texttt{\textbackslash n}Step 1: Review the document information and make four guesses on the topic label.\texttt{\textbackslash n}Step 2: Review the top words and refine each response. \texttt{\textbackslash n}Step 3: Choose the best answer from your guesses and format it like so: <<[ANSWER]>>.\texttt{\textbackslash n}Here is the Document Information:"} with the following features: 3 sampled words from the top abstracts; sampling from the top 4 titles; and sampling 5 words from the top 8 words, produced the best labels with \texttt{Meta-Llama-3-
8B-Instruct} for our knowledge graph dataset. 

\section{Conclusion}
\label{sec:conclusion}


This case study demonstrates that our TopicTag algorithm can effectively generalize in-domain to provide concise label summarizations for document clusters. This is achieved by incorporating the outputs of NMFk as document features within LLM prompts. \texttt{Llama-3-8B-Instruct} exhibited the best performance (with an average SME rating of 3.78 out of a 5-point scale); however, it is noteworthy that smaller models underwent more rounds of optimization. This finding is significant as it suggests that a less computationally intensive and less expensive model can be tuned to outperform state-of-the-art models for this task.

This area of research is promising yet underexplored, particularly in terms of aligning automated optimization processes with stronger NLG metrics. Future work could address this by training an additional LLM to generate task-specific embeddings for labeling, using these embeddings in a manner similar to BERTScore. Alternatively, investigating inter-annotator agreement (IAA) variance could enhance reliability, enabling the application of Reinforcement Learning from Human Feedback (RLHF) for labeling. This approach aligns with other studies highlighting the importance of human preference ratings in achieving superior NLP task performance.

\begin{acks}
  This manuscript has been assigned LA-UR-24-25486. This research was funded by the LANL LDRD grant 20230067DR and the LANL Institutional Computing Program, supported by the U.S. DOE NNSA under Contract No. 89233218CNA000001.
\end{acks}

\bibliographystyle{ACM-Reference-Format}
\bibliography{references.bib}

\end{document}